\documentclass[10pt,twocolumn,letterpaper]{article}

\usepackage{iccv}              

\usepackage{times}
\usepackage{epsfig}
\usepackage{graphicx}
\usepackage{amsmath}
\usepackage{amssymb}

\usepackage{booktabs}
\usepackage{multirow}
\usepackage{caption}
\usepackage{makecell}
\usepackage{algorithm,algpseudocode}
\usepackage{colortbl}
\usepackage[svgnames]{xcolor}
\usepackage{enumitem}
\usepackage{xspace}

\usepackage{stfloats}

\usepackage{tcolorbox} 
\usepackage{listings}

\definecolor{myyellow}{RGB}{255, 255, 242}

\lstdefinestyle{mystyle}{
    basicstyle=\linespread{0.8}\ttfamily\footnotesize,  
    backgroundcolor=\color{myyellow},  
    showstringspaces=false,      
    breaklines=true,
    keywordstyle={},
    commentstyle={},
    breakindent=0pt, 
    breakatwhitespace=true,
}
\definecolor{iccvblue}{rgb}{0.21,0.49,0.74}
\usepackage[pagebackref,breaklinks,colorlinks,allcolors=iccvblue]{hyperref}

\def\paperID{8187} 
\def\confName{ICCV}
\def\confYear{2025}

\title{VTimeCoT: Thinking by Drawing for Video Temporal Grounding and Reasoning}

\author{Jinglei Zhang$^{1,}$\textsuperscript{*}\quad  Yuanfan Guo$^{2,}$\textsuperscript{*}\quad Rolandos Alexandros Potamias$^{3}$ \\
Jiankang Deng$^{3}$\quad Hang Xu$^2$\quad Chao Ma$^{1,}$$^{\dag}$ \\
 $^1$ Shanghai Jiao Tong University\quad $^2$ Noah's Ark Lab\quad $^3$ Imperial College London\\
{\tt\small \{{zhangjinglei168, chaoma}\}@sjtu.edu.cn} \\{\tt\small \{{guoyuanfan1, xu.hang}\}@huawei.com}, \ 
{\tt\small \{{j.deng16, r.potamias}\}@imperial.ac.uk}
}

\newcommand{\shorttitle}{VTimeCoT\xspace}

\begin{document}
\twocolumn[{
\renewcommand\twocolumn[1][]{#1}%
\maketitle
\begin{center}
    \centering
    \captionsetup{type=figure}
    \includegraphics[width=\textwidth]{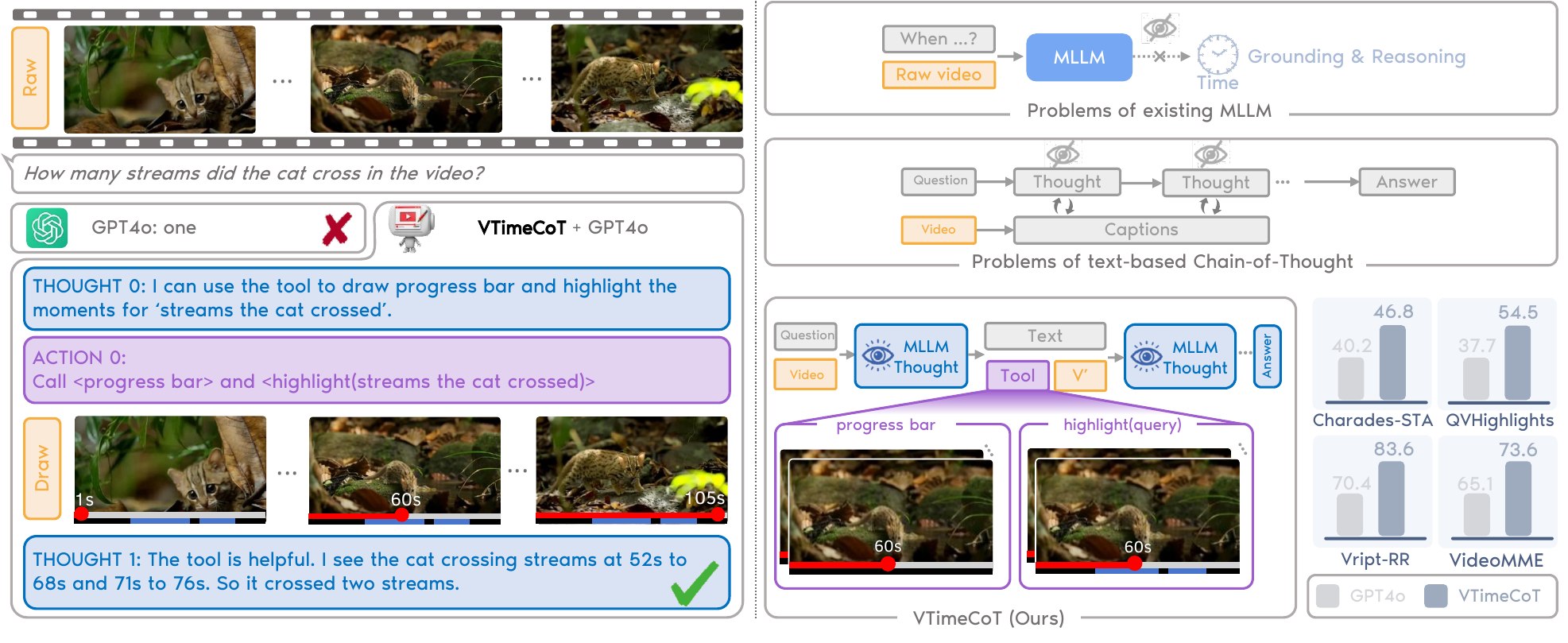}
    \captionof{figure}{We propose \textbf{\shorttitle}, a \textbf{V}isual \textbf{Time} \textbf{C}hain-\textbf{o}f-\textbf{T}hought framework for video temporal grounding and reasoning. \shorttitle constructs cross-modality reasoning across both video and text, which enables the MLLM to utilize progress bar tools to annotate the time progression and highlight the key relevant segments to answer complex temporal questions.}
    \label{fig:teaser}
\end{center}}]

\footnotetext{\textsuperscript{*}Equal contribution. \quad $^{\dag}$Corresponding author.}

\begin{abstract}
In recent years, video question answering based on multimodal large language models (MLLM) has garnered considerable attention, due to the benefits from the substantial advancements in LLMs. However, these models have a notable deficiency in the domains of video temporal grounding and reasoning, posing challenges to the development of effective real-world video understanding systems. Inspired by how humans use video players to interact with the progress bar for video comprehension, we introduce VTimeCoT, a simple yet effective training-free framework, designed for high-performance video grounding and reasoning. The proposed framework incorporates two novel visual tools of the progress bar: a plug-and-play progress bar integration tool and a high-efficiency highlighting tool. In addition, to address the limitations of conventional text-based chain-of-thought (CoT) approaches, we introduce a visuotemporal CoT process that integrates cross-modality reasoning across both video and text. Our approach demonstrates significant performance improvements on both Qwen2VL-7B and GPT4o baselines in tasks of video temporal grounding and reasoning-based question answering. Finally, we showcase that the proposed framework achieves a compositional and interpretable reasoning process. Project page: \href{https://vtimecot.github.io}{https://vtimecot.github.io}. 


\end{abstract}    
\section{Introduction}


Video understanding is a longstanding problem in computer vision and has attracted more attention with the emergence of large language models (LLM)~\cite{hurst2024gpt4o,team2024gemini,wang2024qwen2,chen2024internvl25}. 
Video QA is a representative task reflecting the video reasoning ability.
Given that videos typically contain numerous events occurring at different time points, accurately answering questions is a highly challenging task. Recently, some end-to-end video understanding models based on multimodal large language models (MLLM)~\cite{zhang2023videollama, li2023videochat, lin2023videollava, zhang2024llavavideo, lin2024vila, li2024llavaonevision} have been proposed to address video question-answering tasks, exhibiting remarkable capabilities. However, despite their ability to generate seemingly plausible results, these methods have been shown to exhibit a notable deficiency in temporal grounding~\cite{huang2024vtimellm} and fail to provide a temporal-grounded reasoning process.

With the advent of tool usage capabilities in LLMs, several agent-based video question-answering methods have been proposed ~\cite{min2024morevqa, fan2024videoagent, wang2024videoagent, shang2024traveler,yang2024doraemongpt}, enabling step-by-step video understanding. These methods prompt the LLMs to sequentially reason and invoke external tool models ($e.g.$, object detection and captioning models) to collect textual clues, ultimately inferring the final answer. This, so called, chain-of-thought approach offers significant potential, as it is compositional and training-free, demonstrating impressive zero-shot performance. However, such methods predominantly rely on language and captions as intermediaries, limiting their ability to directly capture the visuotemporal dynamics inherent in videos, especially in the case of long videos. 
Inspired by the time-progress bar commonly used in video players during human interactions, we identify it as an intuitive tool for grounding time and enhancing video comprehension.
In contrast to written language, the progress bar, along with the concurrent visual content, directly conveys the concept of temporal progression.
The video progress bar can represent temporal relationships through sequential positioning or more abstract temporal dependencies. 

In this work, we propose a simple yet effective framework that empowers multimodal LLMs to utilize a video progress bar to construct a \textbf{V}isual \textbf{Time} \textbf{C}hain-\textbf{o}f-\textbf{T}hought (VTimeCoT), facilitating long video temporal grounding and reasoning. 
Inspired by LLMs coupled with visual programming~\cite{gupta2023visual, suris2023vipergpt}, we propose enabling multimodal LLMs to generate code that overlays a progress bar on the video.
First, we construct a frame-sync progress bar integration tool that can be invoked by the MLLM, which generates a progress bar displayed at the bottom of the video, marking the progress and annotating the timestamps on each frame. 
This is naturally adapted to long video and enables the MLLM to perceive the speed of temporal progress directly from the annotated video frames, adapting seamlessly to any frame-per-second (FPS) sampling rate. 
Second, to guide the model's attention on key relevant temporal segments, we propose a training-free and long-video-adaptive highlighting tool for the progress bar based on video-text similarity. 
Specifically, we leverage a robust video-text foundation model to identify the time intervals with the top-k highest similarity and highlight them on the visual progress bar. 
Finally, we prompt the multimodal LLM to invoke the progress bar tools within a visuotemporal chain-of-thought process that integrates text, program, and video annotated with the progress bar to facilitate long video reasoning. At each thought step, the model performs cross-modality reasoning and determines whether to dynamically update the video memory while progressively inferring the answer.
As shown in \cref{fig:teaser}, to determine how many times a cat crosses a stream in the input video, the model first invokes tools to generate a progress bar, identifying the key timestamps corresponding to `streams the cat crossed'. By analyzing the progress bar annotated frames, the model infers the exact timing of each crossing event and accurately determines the number of crossings. This approach enhances the model's temporal reasoning capabilities by seamlessly integrating temporal-grounded visual evidence. 

We demonstrate the effectiveness of the proposed framework across a wide range of video temporal grounding and video-QA tasks. 
The proposed method consistently outperforms the baseline models, Qwen2VL and GPT-4o in temporal grounding tasks, achieving significant performance gains with an average IoU improvement of 6.58\% and 16.83\% over GPT-4o across Charades-STA~\cite{gao2017tall} and QVHighlights~\cite{lei2021detecting} benchmarks. 
Similarly, \shorttitle significantly improves the accuracy of questions related to temporal retrieval, event counting, and event ordering in reasoning-based question-answering tasks. Specifically, the proposed method consistently surpasses state-of-the-art methods across Vript-RR~\cite{yang2024vript} and VideoMME~\cite{fu2024video} benchmarks.

To sum up, we present \shorttitle, a training-free model for temporal reasoning based on the visual progress bar. Specifically:

\begin{itemize}[noitemsep, topsep=0pt, left=0pt, labelsep=0.5em]
    \item We propose the first, to the best of our knowledge, visual time chain-of-thought framework for video temporal grounding and reasoning.  In contrast to previous methods that rely solely on textual reasoning, we leverage the visual progress bar as a medium, making a significant step towards real-world video understanding systems.
    \item The proposed reasoning framework leverages two novel tools from video progress bar integration and highlighting, enabling the MLLM to accurately perceive the timestamp of each frame and identify key temporal segments.
    \item The proposed method enhances the performance of MLLMs by a large margin in video temporal grounding and reasoning tasks through a training-free approach, while also demonstrating strong reasoning capabilities.
\end{itemize}



\section{Related Work}

\noindent \textbf{Multimodal LLM for Video.}
In contrast with traditional video understanding such as obeject detection, segmentation and pose estimation~\cite{yolo, ravi2024sam, potamias2025wilor, zhang2025hawor}, recent multimodal LLMs enable understanding through natural language.
End-to-end video multimodal LLMs are trained on large-scale datasets, to align visual inputs with the language modality and integrate with large language models. Video-ChatGPT~\cite{maaz2023videochatgpt} leveraged CLIP~\cite{radford2021learning} to extract spatial and temporal features from videos and integrate them with Vicuna~\cite{vicuna2023}.
Video-LLaMA~\cite{zhang2023videollama} proposed a video Q-Former designed to enhance the modeling of temporal variations. Subsequently, a series of video multimodal LLM variants~\cite{li2024llama, lin2023videollava, li2024mvbench, xu2024pllava, li2024llavanext, li2024llavaonevision, cheng2024videollama2} have introduced advancements in enhancing video representation, expanding data scale, and refining training methods, leading to improved video question-answering performance. 
However, these end-to-end models operate as black boxes, directly generating answers without an explicit video reasoning process. In contrast, we propose a video understanding framework that incorporates a structured reasoning process.


\noindent \textbf{Programming and Tool Usage of LLM.}
Due to the strong code-generation capabilities of LLMs, several works~\cite{gupta2023visual, suris2023vipergpt, yang2023mmreact, hu2024vpd, choudhury2023zero, liu2024llavaplus} attempted to prompt LLMs to call compositional visual tools to address visual question-answering tasks. VisProg~\cite{gupta2023visual} and ViperGPT~\cite{suris2023vipergpt} defined a variety of image foundation functions (e.g., object detection) and prompted the LLMs to generate code that invokes these functions for answering questions. MM-REACT~\cite{yang2023mmreact} tackled visual tasks through multi-step reasoning guided by text prompts to the LLM, where each step involves thought and calling visual tools. More recently, several studies~\cite{min2024morevqa, yang2024doraemongpt, fan2024videoagent, wang2024videoagent, shang2024traveler} proposed tool-usage frameworks tailored for video understanding. By designing specialized video tools and calling steps, these approaches extract textual cues from videos to facilitate question answering. 
In contrast to the aforementioned works that solely rely on text as the reasoning medium, we introduce a visual progress bar, enabling more effective integration of visual and temporal cues.

\noindent \textbf{Visual Prompt for Multimodal LLM.}
LLMs have been shown to enhance their understanding capabilities by incorporating overlaid visual prompts on images (e.g., circles, keypoints)~\cite{shtedritski2023does, yang2023set, hu2024visual, wu2024dettoolchain, menon2024whiteboard, akter2024self, cai2024vipllava, wu2025mind, li2025imagine}.
SoM~\cite{yang2023set} proposed to utilize SAM~\cite{kirillov2023segment} segmentation model to overlay mask colors and labels on images as input of GPT-4V~\cite{achiam2023gpt}, demonstrating strong zero-shot performance on vision-grounding tasks. Hu \etal~\cite{hu2024visual} introduced a chain-of-thought framework combined with visual prompts to draw markers on images during reasoning, effectively enhancing the image understanding performance of MLLMs. 
However, all previous methods are limited to spatial prompts, limiting their understanding capabilities. In this work, we propose to design the progress bar as visual-temporal prompts to facilitate temporal reasoning.

\section{Method}

\begin{figure*}[ht]
  \centering
  \includegraphics[width=\textwidth]{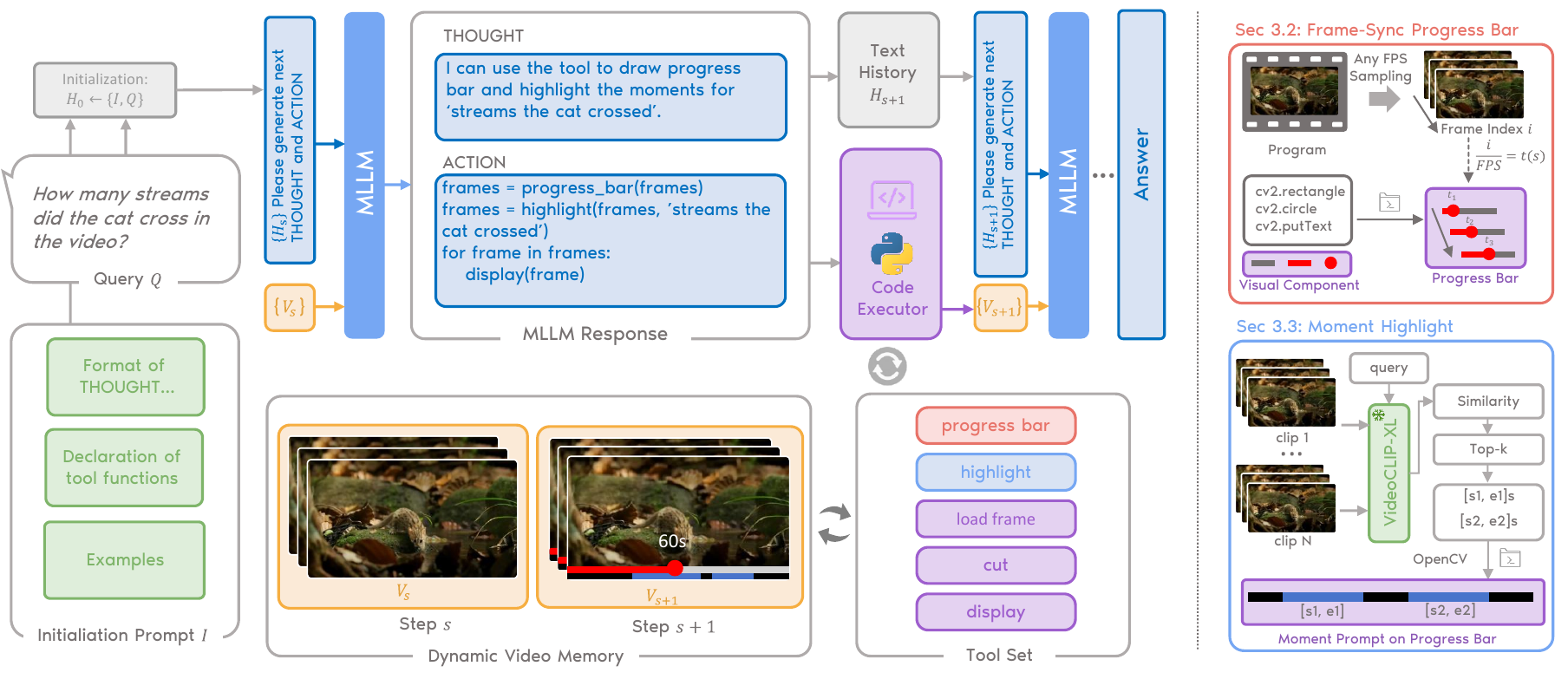}
  \caption{\textbf{Overview of our method.} On the left, we demonstrate how the framework iteratively generates thoughts and actions, which dynamically updates the video memory with an overlaid progress bar for reasoning. On the right, we illustrate two novel tools that integrate the frame-sync visual progress bar and highlight key moments.}
  \label{fig:method}
\end{figure*}

We propose VTimeCoT, a general framework that empowers multimodal large language models using a visual progress bar as an intermediate reasoning step, facilitating temporal grounding and reasoning. 
\cref{fig:method} illustrates how our approach works. Given a raw video and a question, our method generates a chain-of-thought and invokes a set of tools for progress bar integration. 
By utilizing the visual cues from the progress bar as a medium, the model progressively infers the final answer.

\subsection{Visuotemporal Chain-of-Thought}



In video understanding, particularly for long videos, existing MLLM methods can only  provide static answers, which limits the model's ability to focus on critical scenes and results in significant shortcomings in real-world settings. Since the MLLMs fail to identify key scenes within a single forward pass, it is necessary to interact with the video to progressively infer the answer. Therefore, we propose a framework based on interactive mechanisms and dynamic video memory.

Our framework tackles video grounding and reasoning tasks through an iterative interaction with the MLLM, built upon a tool set leveraging the progress bar. Given a textual question and a video as input, VTimeCoT generates a sequence of thoughts and actions to dynamically update the video memory. By manipulating the video, it acquires the necessary information to answer the question.
In this process, the MLLM reasons by plotting and analyzing the progress bar, seamlessly integrating both textual and visual reasoning into a \emph{Visuotemporal Chain-of-Thought}. 
As illustrated in~\cref{fig:method}, VTimeCoT first provides the MLLM with a formatted prompt for initialization and executes actions at each time step \( t \). 
The tool set includes the \textless progress bar\textgreater, \textless highlight\textgreater, \textless cut\textgreater, etc. The \textless cut\textgreater tool is utilized to trim specific segments when the MLLM determines that the video is too long to find the required information. The pseudo-code is presented in~\cref{alg:cot}, where $\mathbb{V}$ and $\mathbb{L}$ represent the modalities of video and language, respectively. 
In \cref{alg:cot}, $p$ serves as a parser to retrieve specific sub-strings by keyword, and $c$ is a code builder that integrates the toolset library and generated code.

\noindent \textbf{Initialization Prompt.} To enable the MLLM to perform step-by-step reasoning and a structured output, we follow ~\cite{hu2024visual} and construct a set of initialization prompts. These prompts define the specific output structure (formatted by keywords of `THOUGHT', `ACTION', and `TERMINATE') for each step and declare the Python tools that the model should invoke. The initialization prompts, along with the question and video frames, are then fed into the MLLM, triggering the iterative loop.

\noindent \textbf{Thought.} During this step, the model analyzes the historical context and video memory to generate its reasoning for the current step. It determines whether a direct answer can be provided directly in this step or if tool assistance is needed, and if so, which tools should be invoked.

\noindent \textbf{Action.} Following the thought step, the model generates an executable Python script utilizing the given tool functions. The framework then builds the code with the toolset library and executes the script to manipulate the video frames.

\noindent \textbf{Dynamic Video Memory.} Using the manipulated video frames we subsequently update the frames in the video memory. The updated context with the manipulated video frames is then fed into the MLLM to proceed with the next step of thought.

To terminate the loop, the model determines at each step whether to cease reasoning and generate the `TERMINATE' keyword to end. To prevent excessively prolonged reasoning, we impose a maximum step limit, enforcing termination at step \( T \).




\begin{algorithm}[t]
\caption{Visuotemporal Chain-of-Thought}
\textbf{Input}: $\{V, Q\|V \in \mathbb{V}, Q \in \mathbb{L}\}; I \in \mathbb{L};$\\
$ \mathtt{MLLM} : \mathbb{V}, \mathbb{L} \rightarrow \mathbb{L}; \text{Toolset}:\{T_{1}, \ldots, T_{n}\} \in \mathbb{L}$
\begin{algorithmic}[1] %
\State $s \gets 0; V_0 \gets V$
\State $H_{0} \gets \{I, Q\}$ \Comment{Initialization prompt and query}
\While{true} 
        \State $y_{s} \gets \mathtt{MLLM}~(V_s, H_{s})$
        \State $\text{THOUGHT}_{s}, \text{ACTION}_{s}, \text{TERMINATE}_{s} \gets p(y_s)$ 
        \If{not $\text{TERMINATE}_{s}$} 
            \State $C_{s} \gets c(\text{ACTION}_{s}, \text{Toolset})$ \Comment{Build code}
            \State $V_{s+1}\gets C_{s}(V_s) $  \Comment{Update video}
            \State $H_{s+1} \gets \{H_s, y_s\}$ \Comment{Update history}
        \Else  
            \State \Return{$y_{s}$} \Comment{Respond final answer}
        \EndIf
        \EndWhile
\end{algorithmic}
\label{alg:cot}
\end{algorithm}

\subsection{Frame-Sync Visual Progress Bar Integration}

Enabling MLLM to perceive the precise timestamp of each frame is a significant challenge and an unresolved problem, due to the diverse temporal sampling rates during MLLM training and inference without time. 
Although recent approaches~\cite{huang2024vtimellm, huang2024lita} have attempted to address this issue by incorporating temporal positional embeddings before the video encoder and introducing temporal-grounding training tasks, their flexibility is limited due to the scarcity of specific data and the need for fine-tuning. This issue is particularly pronounced in long videos, where the token limitations of MLLMs prevent them from accurately perceiving the timestamp of each frame.

Inspired by visual programming paradigms~\cite{gupta2023visual,suris2023vipergpt}, we propose a simple yet effective method to empower MLLMs to perceive the precise timestamp of each frame, by generating the video progress bar. 
Leveraging the code generation capabilities of MLLMs, our approach prompts the MLLM to invoke Python plotting tools and use their universal OCR and shape understanding capabilities to perceive time, without any additional training requirements.
Specifically, given a raw video $V \in \mathbb{R}^{T \times H \times W \times 3}$ as input, this tool can generate the progress bar at the bottom of the original frames and return the annotated frames.
To enable adaptation to arbitrary FPS sampling, we perform a frame-synchronization step to convert frame indexes to the real second time, as shown in~\cref{fig:method}. To enable MLLM to perceive seconds directly without any additional hour-minute-second conversions, we design the timestamps using a straightforward seconds format.
To construct a robust visual prompt that can be easily interpreted from the MLLM, we employ compositional components using elongated rectangles, circles, and timestamps to plot the progress bar.
Finally, the progress bar is placed at the bottom of raw frames, which can be expressed as:
\begin{equation}
    V'_t =  V_t \oplus T_t,
\end{equation}
where \( \oplus \) represents the vertical concatenation operation, \( T_t \) denotes the image of the progress bar generated by OpenCV at time \( t \) and \( V'_t \) is the resulting frame. We wrap this integration process into a Python tool function, allowing MLLM to invoke it for generating the annotated video. 

\subsection{Moment Highlight by Video-Text Similarity}


Although multimodal LLMs have made significant progress in video content captioning, recent studies~\cite{huang2024vtimellm, huang2024lita} have revealed substantial deficiencies in their temporal grounding capabilities.
Existing works~\cite{huang2024vtimellm, huang2024lita} rely on constructing temporal grounding instructional data and require finetuning to emerge such capability. Besides, due to token limitations in MLLMs, they struggle to perform temporal grounding for long videos.
In contrast, we propose a training-free and long-video-adaptive method to enhance the temporal grounding capabilities of MLLMs, leveraging the similarities from the robust video-text foundation models.

Given a query and a video $V \in \mathbb{R}^{T \times H \times W \times 3}$ containing multiple events, our proposed tool estimates the query-relevant temporal segments, including their start and end times, and generates a highlighted visual prompt.

\noindent \textbf{Moment Retrieval by Video-Text Similarity.} To construct a robust zero-shot moment retrieval module, capable of generalizing across a wide range of videos, we employ the VideoCLIP-XL~\cite{wang2024videoclip} foundation model as our embedding extractor. 
The video is initially processed at a high temporal resolution of $r$ FPS. We group every 8 frames into a video clip, resulting in $N$ clips. 
It should be noted that since the temporal retrieval module operates as an external component, it can efficiently handle high frame rates without being constrained by the token count of the MLLM. This makes our method adaptive to long-form video.
These clips are then processed by the VideoCLIP-XL visual encoder, producing $N$ video embeddings. 
We use VideoCLIP-XL text encoder to extract the text embedding for the text query and we compute the cosine similarity between the query embedding and the video embeddings of each clip:
\begin{equation}
    \text{sim}(i) = \frac{e^{v}_i \cdot e^q}{|e^{v}_i| |e^q|},
\end{equation}
where $e^{v}_i$ denotes the embedding of the $i$-th clip and $e^q$ is the query embedding. Using top-$k$ selection, we obtain the $k$ clips with the highest similarity and extract the start and end timestamps of each contiguous segment.


\noindent \textbf{Moment Highlight on the Progress Bar.} 
Motivated by a series of works that demonstrated the effectiveness of visual cues as a medium for reasoning beyond text~\cite{hu2024visual, menon2024whiteboard, akter2024self, wu2025mind, li2025imagine}, we propose using moment-highlighting cues for video temporal reasoning. 
As shown in~\cref{fig:method}, we construct the highlight tool using OpenCV library to plot colorful highlight masks on specific intervals of the progress bar. This tool takes the video frames and time intervals as input and annotates the highlighted progress bar under the frames. To ensure that the MLLM can easily interpret the highlights, we utilize different colors between the mask and the progress bar. We wrap the aforementioned retrieval and highlighting process as a Python tool function, enabling the MLLM to invoke it to manipulate the input video frames.

\section{Experiments}


In this section, we first present the implementation details, datasets and evaluation metrics. Subsequently, we assess the performance of the proposed framework on the video temporal grounding benchmarks (\cref{sec:grounding}). To further evaluate the reasoning capabilities, we conducted extensive quantitative and qualitative analyses on the ``reasoning based on retrieving" benchmark of Vript-RR (\cref{sec:rr}). In addition, we demonstrate the significant role our reasoning framework plays in enhancing long video question-answering performance (\cref{sec:vqa}). Finally, we give the ablations of key modules (\cref{sec:ablation}).

\noindent\textbf{Implementation Details.} In the framework, we conducted experiments on two core MLLMs, Qwen2VL-7B~\cite{wang2024qwen2} and GPT4o-20240513~\cite{hurst2024gpt4o}, as they represent the state-of-the-art open-source and closed-source MLLM models, respectively. We set the decoding temperature of the MLLM to 0. By default, the input for the MLLM consists of 32 frames uniformly sampled from the video. For video frames, $H$ and $W$ are resized to make the longer side 480 pixels. Our agent implementation is based on AutoGen~\cite{wu2023autogen}. The implementations of tool functions are adapted from Gradio~\cite{abid2019gradio}.
For GPT4o, we utilize the OpenAI API service. 
The details of the MLLM prompts are provided in the supplementary materials.
For VideoCLIP-XL we use FPS $r=1$ to sample video frames and group every 8 frames as a clip to extract embedding. Regarding the top-$k$ selection of clips, $k$ is empirically set to 8. The maximum number of reasoning steps $T$ is set to 3.

\noindent \textbf{Datasets and Evaluation Metrics.}
To evaluate the proposed VTimeCoT method, we utilized four benchmark datasets, spanning a wide range of video temporal grounding, reasoning, and long-video question-answering tasks.

\noindent \textbf{Charades-STA}~\cite{gao2017tall} is a benchmark dataset for temporal grounding, composed of 1334 videos along with the corresponding start-end frame annotations of 3720 queries. The average video length is 30 seconds. We follow ~\cite{huang2024vtimellm} and report mean IoU (mIoU) and recall@1, IoU $\geq m$(R@$m$) metrics, where $m$ includes 0.3, 0.5 and 0.7. IoU denotes the intersection over the union between the predicted and ground truth time segments. 

\begin{table}[t]
\centering
\small{
\setlength{\tabcolsep}{1pt}
\begin{tabular}{lcccc}
\toprule 
Method & R1@0.3 & R1@0.5 & R1@0.7 & mIoU \\
\midrule  
 VideoChat-7B~\cite{li2023videochat} & 9.00 & 3.30 & 1.30 & 6.50 \\
 VideoLLaMA-7B~\cite{zhang2023videollama} & 10.40 & 3.80 & 0.90 & 7.10 \\
 VideoChatGPT-7B~\cite{maaz2023videochatgpt} & 20.00 & 7.70 & 1.70 & 13.70 \\
 LLAVA-Onevision-7B~\cite{li2024llavaonevision}  & 33.04 & 11.05 & 4.11 & 20.98 \\
 VTimeLLM-7B~\cite{huang2024vtimellm}  & 51.00 & 27.50 & 11.40 & 31.20 \\
 VTimeLLM-13B~\cite{huang2024vtimellm}  & 55.30 & 34.30 & 14.70 & 34.60 \\
 \midrule
 Qwen2VL-7B~\cite{wang2024qwen2} & 37.31 & 12.85 & 4.11 & 24.34  \\
  $\text{\shorttitle}_{\text{Qwen2VL-7B}}$ & \textbf{66.96} & \textbf{38.79} & \textbf{20.83} & \textbf{43.41} \\
 \midrule 
 GPT4o~\cite{hurst2024gpt4o}  & 63.76 & 37.12 & 14.65 & 40.20  \\
 $\text{\shorttitle}_{\text{GPT4o}}$  & \textbf{74.06} & \textbf{51.02} & \textbf{22.45} & \textbf{46.78} \\
\bottomrule
\end{tabular}
\caption{\textbf{Quantitative comparison on the Charades-STA dataset for temporal grounding.} We report the mIoU and the recall performance at different IoU thresholds.  
}
\label{tab:charades_sta}}
\end{table}

\begin{table}[t]
\centering
\small{
\setlength{\tabcolsep}{1pt}
\begin{tabular}{lcccc}
\toprule 
Method & R1@0.3 & R1@0.5 & R1@0.7 & mIoU \\
\midrule  
 LLAVA-Onevision-7B~\cite{li2024llavaonevision}  & 34.91 & 17.91 & 9.7 & 25.53 \\
 VTimeLLM-7B~\cite{huang2024vtimellm} & 44.58 & 25.03 & 9.29 & 28.99 \\
 \midrule 
 Qwen2VL-7B~\cite{wang2024qwen2} & 31.87 & 14.65 & 7.35 & 22.77  \\
 $\text{\shorttitle}_{\text{Qwen2VL-7B}}$& \textbf{67.50} & \textbf{45.79} & \textbf{25.11} & \textbf{46.21} \\
 \midrule
 GPT4o~\cite{hurst2024gpt4o}  & 55.61 & 35.68 & 19.29 & 37.66  \\
 $\text{\shorttitle}_{\text{GPT4o}}$  & \textbf{79.35} & \textbf{59.74} & \textbf{33.81} & \textbf{54.49} \\
\bottomrule
\end{tabular}
\caption{\textbf{Quantitative comparison on the QVHighlights dataset for temporal grounding of discontinuous segments.} We report the mIoU and the recall performance at different IoU thresholds.}
\label{tab:qvhighlights}}
\end{table}

\noindent \textbf{QVHighlights}~\cite{lei2021detecting} is a benchmark dataset for semantic relevance and temporal grounding from discontinuous time segments.
It contains 1519 videos, with an average length of 150 seconds, and 1550 queries with time annotations, where each query spans multiple discontinuous time segments within a video. For evaluation, we report the commonly used mIoU and R@$m$ metrics.

\noindent \textbf{Vript-RR}~\cite{yang2024vript} is a challenging benchmark dataset for scene retrieval and multi-hop reasoning. 
The average length of the videos is 622 seconds. It contains 152 questions, each accompanied by a hint to locate the scene that the question refers to. The benchmark includes both multiple-choice and open-ended question-answering settings. We follow ~\cite{yang2024vript} and evaluate the open-ended accuracy using GPT4.

\noindent \textbf{VideoMME}~\cite{fu2024video} is a video question-answering benchmark tailored for MLLMs, featuring diverse videos ranging from 11 seconds to 1 hour in length. It contains 900 videos and 2700 question-answer pairs. We follow ~\cite{fu2024video} to report accuracy metrics under two evaluation settings with subtitles and without subtitles.

\subsection{Temporal Grounding}
\label{sec:grounding}


To ensure accurate temporal reasoning, it is essential to achieve robust and high-performance temporal grounding. 
In ~\cref{tab:charades_sta}, we compare \shorttitle with two state-of-the-art baselines, Qwen2VL-7B and GPT4o, to identify the start and end timestamps of the query event. 
Specifically, using both Qwen2VL-7B and GPT4o models as the core MLLM of our framework, we evaluate the performance of \shorttitle to localize the query event. 
As can be easily observed, although these advanced MLLMs achieve state-of-the-art performance in video content recognition, they fall short in accurately grounding temporal boundaries for events in videos, facing difficulties in perceiving the precise timestamp and the boundaries of events. In contrast, the proposed method effectively addresses these challenges, providing accurate visual cues of the progress bar through the reasoning process that facilitates the grounding performance in videos. 
Note that \shorttitle demonstrates superior performance without any additional training cost, even compared to the VTimeLLM model, which has been meticulously fine-tuned on extensive temporal grounding data.

\begin{table}[t]
\centering
\small{
\begin{tabular}{lcc}
\toprule 
Methods  & Multi-Choice & Open\\
\midrule  
 VideoChatGPT~\cite{maaz2023videochatgpt} & 29.60 & 17.80 \\
 Video-LLaMA~\cite{zhang2023videollama} & 28.30 & 14.50 \\
 VideoChat~\cite{li2023videochat} & 22.40 & 15.10  \\
 VideoChat2~\cite{li2024mvbench} & 42.10 & 22.40  \\
 ST-LLM~\cite{liu2024st} & 33.60 & 26.30 \\
 PLLaVA 7B~\cite{xu2024pllava} & 55.30 & 36.20  \\
 VILA-1.5 8B~\cite{lin2024vila} & 55.30 & 32.30 \\
 \midrule 
 Qwen2VL-7B~\cite{wang2024qwen2} & 59.87 & 35.95  \\ 
 $\text{\shorttitle}_{\text{Qwen2VL-7B}}$& \textbf{62.50} & \textbf{41.45} \\
 \midrule 
 GPT4o~\cite{hurst2024gpt4o}  & 70.39 & 61.18 \\
 $\text{\shorttitle}_{\text{GPT4o}}$  & \textbf{83.55} & \textbf{68.42}\\
\bottomrule
\end{tabular}
\caption{\textbf{Quantitative comparison on the Vript-RR dataset.} We report the response accuracy on both multiple-choice and open-ended question settings.}
\label{tab:vript_rr}}
\end{table}


To further evaluate the temporal grounding performance of the proposed and the baseline methods in the presence of multiple discontinuous time segments for each query, we utilized the QVHighlights benchmark dataset (\cref{tab:qvhighlights}).  
Given a query and a video, we request from each model to identify start-end timestamps of all temporal segments in the video that fit the query. 
As can be observed, \shorttitle consistently enhances the performance compared to different MLLMs.
On the contrary, baseline MLLM methods struggle with discontinuous time segments, failing to identify multiple temporal spans and leading to degraded grounding performance.

\subsection{Long-Video Reasoning Based on Retrieving}
\label{sec:rr}

\begin{figure*}[t]
  \centering
  \includegraphics[width=\textwidth]{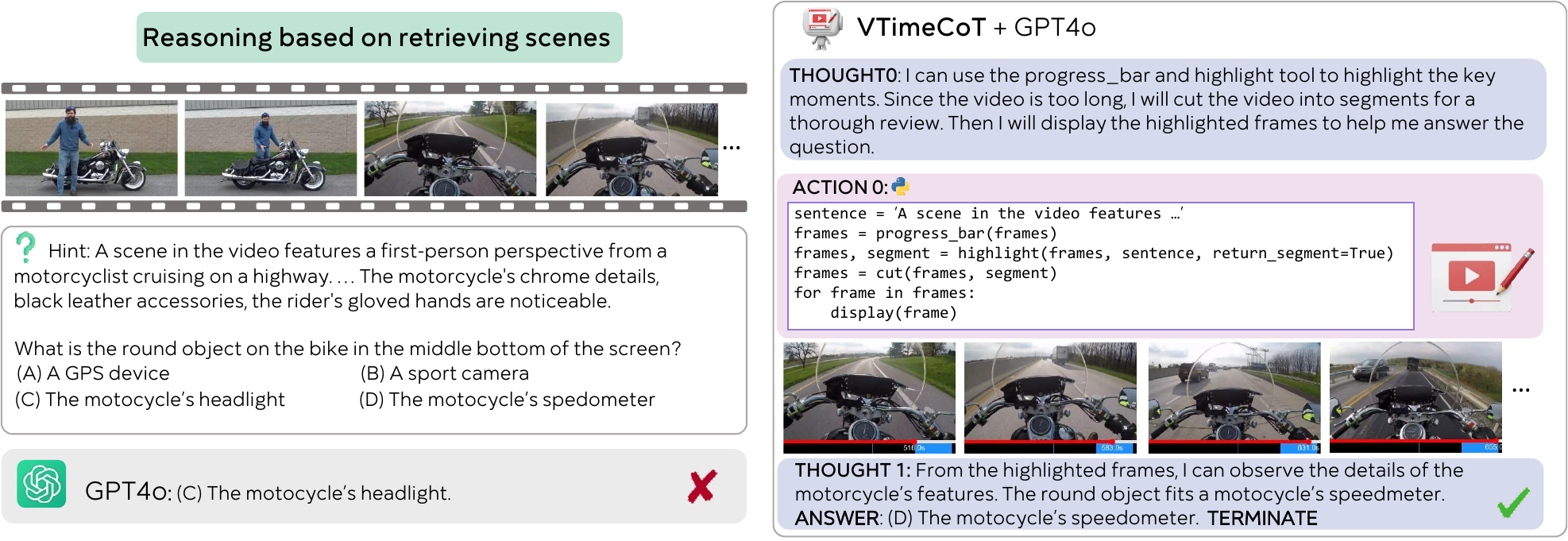}
  \caption{\textbf{Qualitative comparison on Vript-RR benchmark.} 
  In contrast to GPT-4o which is misled by the irrelevant scenes, \shorttitle reasons using the progress bar and highlights the key moments, leading to accurate answers.}
  \label{fig:vriptrr_qualitative}
\end{figure*}

\begin{figure*}[t]
  \centering
  \includegraphics[width=\textwidth]{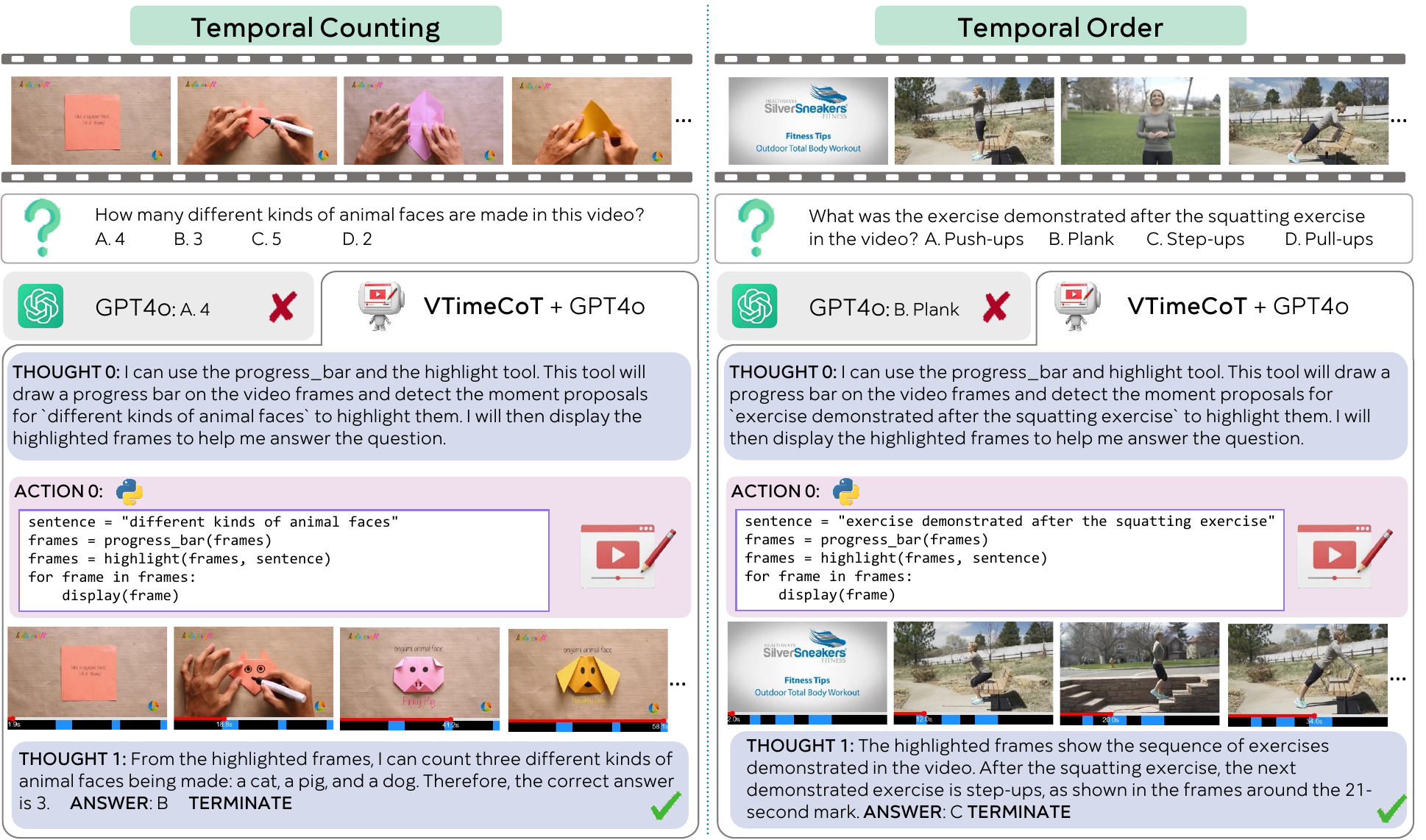}
  \caption{\textbf{Qualitative comparison on VideoMME benchmark.} Through reasoning that integrates the progress bar and highlights, \shorttitle accurately answers questions related to temporal counting and order, which GPT-4o falls short.}
  \label{fig:videomme_qualitative}
\end{figure*}

In~\cref{tab:vript_rr}, we evaluate the question-answering performance of the proposed and the baseline methods on the Vript-RR benchmark, which contains questions that require scene retrieval. 
Given a scene hint and a question, the model must identify the specific scene to deliver an accurate response.
We compared the proposed method with the state-of-the-art baseline methods including Qwen2VL-7B and GPT4o. As can be observed \shorttitle achieves a significant performance improvement, surpassing the baseline methods in both multiple-choice and open-ended settings. 

To further illustrate the superiority of the proposed method in the task of scene retrieval and reasoning, we qualitatively compare GPT4o and \shorttitle on the Vript-RR dataset in \cref{fig:vriptrr_qualitative}. Through multi-step reasoning that leverages the progress bar and highlights the key time interval, \shorttitle accurately identifies the correct answer, while GPT-4o falls short.


\begin{table}[t]
\centering
\small{
\begin{tabular}{lccc}
\toprule 
\multirow{2}{*}{Methods} & \multirow{2}{*}{Frames} &  \multicolumn{2}{c}{Accuracy} \\
 & & w/o subs & w. subs \\
 \midrule  
 \rowcolor{gray!20}
 Gemini-1.5-Flash~\cite{team2024gemini} & 1 fps & 70.3 & 75.0 \\
  \rowcolor{gray!20}
 GPT4o~\cite{hurst2024gpt4o} & 384 & 71.9 & 77.2 \\
  \rowcolor{gray!20}
 Gemini-1.5-Pro~\cite{team2024gemini} & 1 fps & 75.0 & 81.3 \\
 \midrule  
 Video-LLaVA-7B~\cite{lin2023videollava} & 8 & 39.9 & 41.6 \\
 VideoLLaMA2-7B~\cite{cheng2024videollama2} & 16 & 54.9 & 56.4 \\
 LLAVA-Onevision-7B~\cite{li2024llavaonevision} & 32 & 58.2 & 61.5 \\
 LongVILA-7B~\cite{xue2024longvila} & 256 & 60.1 & 65.1 \\
 InternVL2.5-8B~\cite{chen2024internvl25} & 64 & 64.2 & 66.9 \\
 mPLUG-Owl3-7B~\cite{ye2024mplugowl3} & 128 & 59.3 & 68.1 \\
 LLaVA-Video-7B~\cite{zhang2024llavavideo} & 64 & 63.3 & 69.7 \\
 NVILA-8B~\cite{liu2024nvila} & 256 & 64.2 & 70.0 \\
 VideoLLaMA3-7B~\cite{zhang2025videollama3} & 180 & 66.2 & 70.3 \\
\midrule  
 GPT4o ~\cite{hurst2024gpt4o} & 32 & 61.6 & 65.1 \\
 $\text{\shorttitle}_{\text{GPT4o}}$ & 32 & 64.2 & 73.6 \\
\bottomrule
\end{tabular}
\caption{\textbf{Quantitative comparison on the VideoMME dataset.} We report the response accuracy under two settings: with subtitles and without subtitles. 
}
\label{tab:videomme}}
\end{table}

\subsection{Long-Video Question Answering}
\label{sec:vqa}

A pivotal challenge in long-form video question answering lies in accurately inferring the frequency and sequential relationships of temporal events. 
Currently, end-to-end video MLLMs struggle to accurately localize events and capture their temporal dependencies across multiple frames, limiting their ability to generalize to real-world temporal understanding.
In ~\cref{tab:videomme}, we evaluate the performance of the proposed VTimeCoT model in question answering on long videos of the VideoMME benchmark dataset. 
Due to the prohibitive evaluation costs and to ensure a fair comparison, we re-evaluate GPT-4o using 32 frames.
Utilizing robust temporal grounding and reasoning, \shorttitle significantly outperforms the baseline methods in long-video question answering.

\begin{table}[t]
\centering
\small{
\setlength{\tabcolsep}{3pt}
\begin{tabular}{lccccc}
\toprule 
MLLM & CoT & Progress Bar & Highlight & \makecell{QVHighlights\\ (mIoU)}   & \makecell{Vript-RR\\ (M-Acc)} \\
 \midrule 
 GPT4o & $\times$ & $\times$ & $\times$ &  37.66 & 70.39 \\
 GPT4o & \checkmark & $\times$ & $\times$ & 41.85 &  73.68  \\
 GPT4o & \checkmark & \checkmark & $\times$ & 49.40 & 76.32  \\
 GPT4o & \checkmark & \checkmark & \checkmark & 54.49 & 83.55   \\
\bottomrule
\end{tabular}
\caption{\textbf{Ablations on the key modules.} M-Acc is the multi-choice accuracy on Vript-RR.}
\label{tab:ablation}}
\end{table}

The effectiveness of the proposed method can be further validated in \cref{fig:videomme_qualitative}, where we compared the responses of the state-of-the-art GPT4o and \shorttitle. 
In contrast to GPT4o, \shorttitle can accurately identify the correct answers through its systematic, step-by-step reasoning process, even in challenging scenarios involving temporal counting and order discernment. 
It is also worth mentioning that \shorttitle, apart from its superior performance in question-answering accuracy, demonstrates logical and interpretable reasoning steps.

\subsection{Ablations}
\label{sec:ablation}

To further investigate the contribution of each component in the proposed method, we conducted an ablation study. In~\cref{tab:ablation}, we evaluate the importance of the Chain-of-Thought (CoT), progress bar, and highlighting modules. 
As can be observed the GPT4o employing standard CoT without the use of the progress bar ($i.e.$, solely on text-based step-by-step reasoning),  results in significant performance degradation compared to the proposed method. 
To assess the effect of the proposed modules of the progress bar, we developed a configuration that utilizes only the progress bar tool without using the highlighting module. 
For completion, we also report the full VTimeCoT results, utilizing both the progress bar and highlighting tools. 
Both tools contribute to the performance improvement of the proposed method, which highlights the effectiveness of explicitly visualizing the precise timestamp and highlighting the relevant moments.




\section{Conclusion}

In this work, we propose the first visual time framework, designed to formulate a visuotemporal chain-of-thought for video temporal grounding and reasoning. We introduce a plug-and-play tool of the progress bar to generate visual temporal cues from any video, that can enable multimodal large language models (MLLMs) to leverage tool-usage capabilities for accurately perceiving the speed of temporal progression. 
In addition, we propose a zero-shot temporal retrieval tool, built on top of a strong video-text foundational model, augmenting it with temporal grounding capabilities without any further training requirements. We integrate these two tools into a visuotemporal chain-of-thought framework to facilitate cross-modal reasoning between video and text. Through extensive experiments, we demonstrate that the proposed method surpasses previous state-of-the-art baselines in temporal grounding and reasoning-based question-answering benchmarks, demonstrating logical and interpretable reasoning steps. 

\noindent\textbf{Acknowledgments.} This work was supported by NSFC (62322113, 62376156), Shanghai Municipal Science and Technology Major Project (2021SHZDZX0102), and the Fundamental Research Funds for the Central Universities.



{\small
\bibliographystyle{ieeenat_fullname}
\bibliography{main}

\begin{thebibliography}{58}
\providecommand{\natexlab}[1]{#1}
\providecommand{\url}[1]{\texttt{#1}}
\expandafter\ifx\csname urlstyle\endcsname\relax
  \providecommand{\doi}[1]{doi: #1}\else
  \providecommand{\doi}{doi: \begingroup \urlstyle{rm}\Url}\fi

\bibitem[Abid et~al.(2019)Abid, Abdalla, Abid, Khan, Alfozan, and Zou]{abid2019gradio}
Abubakar Abid, Ali Abdalla, Ali Abid, Dawood Khan, Abdulrahman Alfozan, and James Zou.
\newblock Gradio: Hassle-free sharing and testing of ml models in the wild.
\newblock \emph{arXiv preprint arXiv:1906.02569}, 2019.

\bibitem[Achiam et~al.(2023)Achiam, Adler, Agarwal, Ahmad, Akkaya, Aleman, Almeida, Altenschmidt, Altman, Anadkat, et~al.]{achiam2023gpt}
Josh Achiam, Steven Adler, Sandhini Agarwal, Lama Ahmad, Ilge Akkaya, Florencia~Leoni Aleman, Diogo Almeida, Janko Altenschmidt, Sam Altman, Shyamal Anadkat, et~al.
\newblock Gpt-4 technical report.
\newblock \emph{arXiv preprint arXiv:2303.08774}, 2023.

\bibitem[Akter et~al.(2024)Akter, Madaan, Lee, Yang, and Nyberg]{akter2024self}
Syeda~Nahida Akter, Aman Madaan, Sangwu Lee, Yiming Yang, and Eric Nyberg.
\newblock Self-imagine: Effective unimodal reasoning with multimodal models using self-imagination.
\newblock \emph{arXiv preprint arXiv:2401.08025}, 2024.

\bibitem[Cai et~al.(2024)Cai, Liu, Mustikovela, Meyer, Chai, Park, and Lee]{cai2024vipllava}
Mu Cai, Haotian Liu, Siva~Karthik Mustikovela, Gregory~P Meyer, Yuning Chai, Dennis Park, and Yong~Jae Lee.
\newblock Vip-llava: Making large multimodal models understand arbitrary visual prompts.
\newblock In \emph{Proceedings of the IEEE/CVF Conference on Computer Vision and Pattern Recognition}, pages 12914--12923, 2024.

\bibitem[Chen et~al.(2024)Chen, Wang, Cao, Liu, Gao, Cui, Zhu, Ye, Tian, Liu, et~al.]{chen2024internvl25}
Zhe Chen, Weiyun Wang, Yue Cao, Yangzhou Liu, Zhangwei Gao, Erfei Cui, Jinguo Zhu, Shenglong Ye, Hao Tian, Zhaoyang Liu, et~al.
\newblock Expanding performance boundaries of open-source multimodal models with model, data, and test-time scaling.
\newblock \emph{arXiv preprint arXiv:2412.05271}, 2024.

\bibitem[Cheng et~al.(2024)Cheng, Leng, Zhang, Xin, Li, Chen, Zhu, Zhang, Luo, Zhao, et~al.]{cheng2024videollama2}
Zesen Cheng, Sicong Leng, Hang Zhang, Yifei Xin, Xin Li, Guanzheng Chen, Yongxin Zhu, Wenqi Zhang, Ziyang Luo, Deli Zhao, et~al.
\newblock Videollama 2: Advancing spatial-temporal modeling and audio understanding in video-llms.
\newblock \emph{arXiv preprint arXiv:2406.07476}, 2024.

\bibitem[Chiang et~al.(2023)Chiang, Li, Lin, Sheng, Wu, Zhang, Zheng, Zhuang, Zhuang, Gonzalez, Stoica, and Xing]{vicuna2023}
Wei-Lin Chiang, Zhuohan Li, Zi Lin, Ying Sheng, Zhanghao Wu, Hao Zhang, Lianmin Zheng, Siyuan Zhuang, Yonghao Zhuang, Joseph~E. Gonzalez, Ion Stoica, and Eric~P. Xing.
\newblock Vicuna: An open-source chatbot impressing gpt-4 with 90\%* chatgpt quality, 2023.

\bibitem[Choudhury et~al.(2023)Choudhury, Niinuma, Kitani, and Jeni]{choudhury2023zero}
Rohan Choudhury, Koichiro Niinuma, Kris~M Kitani, and L{\'a}szl{\'o}~A Jeni.
\newblock Zero-shot video question answering with procedural programs.
\newblock \emph{arXiv preprint arXiv:2312.00937}, 2023.

\bibitem[Fan et~al.(2024)Fan, Ma, Wu, Du, Li, Gao, and Li]{fan2024videoagent}
Yue Fan, Xiaojian Ma, Rujie Wu, Yuntao Du, Jiaqi Li, Zhi Gao, and Qing Li.
\newblock Videoagent: A memory-augmented multimodal agent for video understanding.
\newblock In \emph{European Conference on Computer Vision}, pages 75--92. Springer, 2024.

\bibitem[Fu et~al.(2024)Fu, Dai, Luo, Li, Ren, Zhang, Wang, Zhou, Shen, Zhang, et~al.]{fu2024video}
Chaoyou Fu, Yuhan Dai, Yongdong Luo, Lei Li, Shuhuai Ren, Renrui Zhang, Zihan Wang, Chenyu Zhou, Yunhang Shen, Mengdan Zhang, et~al.
\newblock Video-mme: The first-ever comprehensive evaluation benchmark of multi-modal llms in video analysis.
\newblock \emph{arXiv preprint arXiv:2405.21075}, 2024.

\bibitem[Gao et~al.(2017)Gao, Sun, Yang, and Nevatia]{gao2017tall}
Jiyang Gao, Chen Sun, Zhenheng Yang, and Ram Nevatia.
\newblock Tall: Temporal activity localization via language query.
\newblock In \emph{Proceedings of the IEEE international conference on computer vision}, pages 5267--5275, 2017.

\bibitem[Gupta and Kembhavi(2023)]{gupta2023visual}
Tanmay Gupta and Aniruddha Kembhavi.
\newblock Visual programming: Compositional visual reasoning without training.
\newblock In \emph{Proceedings of the IEEE/CVF Conference on Computer Vision and Pattern Recognition}, pages 14953--14962, 2023.

\bibitem[Hu et~al.(2024{\natexlab{a}})Hu, Shi, Fu, Roth, Ostendorf, Zettlemoyer, Smith, and Krishna]{hu2024visual}
Yushi Hu, Weijia Shi, Xingyu Fu, Dan Roth, Mari Ostendorf, Luke Zettlemoyer, Noah~A Smith, and Ranjay Krishna.
\newblock Visual sketchpad: Sketching as a visual chain of thought for multimodal language models.
\newblock \emph{arXiv preprint arXiv:2406.09403}, 2024{\natexlab{a}}.

\bibitem[Hu et~al.(2024{\natexlab{b}})Hu, Stretcu, Lu, Viswanathan, Hata, Luo, Krishna, and Fuxman]{hu2024vpd}
Yushi Hu, Otilia Stretcu, Chun-Ta Lu, Krishnamurthy Viswanathan, Kenji Hata, Enming Luo, Ranjay Krishna, and Ariel Fuxman.
\newblock Visual program distillation: Distilling tools and programmatic reasoning into vision-language models.
\newblock In \emph{Proceedings of the IEEE/CVF Conference on Computer Vision and Pattern Recognition}, pages 9590--9601, 2024{\natexlab{b}}.

\bibitem[Huang et~al.(2024{\natexlab{a}})Huang, Wang, Chen, Song, and Zhu]{huang2024vtimellm}
Bin Huang, Xin Wang, Hong Chen, Zihan Song, and Wenwu Zhu.
\newblock Vtimellm: Empower llm to grasp video moments.
\newblock In \emph{Proceedings of the IEEE/CVF Conference on Computer Vision and Pattern Recognition}, pages 14271--14280, 2024{\natexlab{a}}.

\bibitem[Huang et~al.(2024{\natexlab{b}})Huang, Liao, Radhakrishnan, Yin, Molchanov, Yu, and Kautz]{huang2024lita}
De-An Huang, Shijia Liao, Subhashree Radhakrishnan, Hongxu Yin, Pavlo Molchanov, Zhiding Yu, and Jan Kautz.
\newblock Lita: Language instructed temporal-localization assistant.
\newblock In \emph{European Conference on Computer Vision}, pages 202--218. Springer, 2024{\natexlab{b}}.

\bibitem[Hurst et~al.(2024)Hurst, Lerer, Goucher, Perelman, Ramesh, Clark, Ostrow, Welihinda, Hayes, Radford, et~al.]{hurst2024gpt4o}
Aaron Hurst, Adam Lerer, Adam~P Goucher, Adam Perelman, Aditya Ramesh, Aidan Clark, AJ Ostrow, Akila Welihinda, Alan Hayes, Alec Radford, et~al.
\newblock Gpt-4o system card.
\newblock \emph{arXiv preprint arXiv:2410.21276}, 2024.

\bibitem[Kirillov et~al.(2023)Kirillov, Mintun, Ravi, Mao, Rolland, Gustafson, Xiao, Whitehead, Berg, Lo, et~al.]{kirillov2023segment}
Alexander Kirillov, Eric Mintun, Nikhila Ravi, Hanzi Mao, Chloe Rolland, Laura Gustafson, Tete Xiao, Spencer Whitehead, Alexander~C Berg, Wan-Yen Lo, et~al.
\newblock Segment anything.
\newblock In \emph{Proceedings of the IEEE/CVF International Conference on Computer Vision}, pages 4015--4026, 2023.

\bibitem[Lei et~al.(2021)Lei, Berg, and Bansal]{lei2021detecting}
Jie Lei, Tamara~L Berg, and Mohit Bansal.
\newblock Detecting moments and highlights in videos via natural language queries.
\newblock \emph{Advances in Neural Information Processing Systems}, 34:\penalty0 11846--11858, 2021.

\bibitem[Li et~al.(2024{\natexlab{a}})Li, Zhang, Guo, Zhang, Li, Zhang, Zhang, Zhang, Li, Liu, et~al.]{li2024llavaonevision}
Bo Li, Yuanhan Zhang, Dong Guo, Renrui Zhang, Feng Li, Hao Zhang, Kaichen Zhang, Peiyuan Zhang, Yanwei Li, Ziwei Liu, et~al.
\newblock Llava-onevision: Easy visual task transfer.
\newblock \emph{arXiv preprint arXiv:2408.03326}, 2024{\natexlab{a}}.

\bibitem[Li et~al.(2025)Li, Wu, Zhang, Xia, Mao, Dong, Vuli{\'c}, and Wei]{li2025imagine}
Chengzu Li, Wenshan Wu, Huanyu Zhang, Yan Xia, Shaoguang Mao, Li Dong, Ivan Vuli{\'c}, and Furu Wei.
\newblock Imagine while reasoning in space: Multimodal visualization-of-thought.
\newblock \emph{arXiv preprint arXiv:2501.07542}, 2025.

\bibitem[Li et~al.(2024{\natexlab{b}})Li, Zhang, Zhang, Zhang, Li, Li, Ma, and Li]{li2024llavanext}
Feng Li, Renrui Zhang, Hao Zhang, Yuanhan Zhang, Bo Li, Wei Li, Zejun Ma, and Chunyuan Li.
\newblock Llava-next-interleave: Tackling multi-image, video, and 3d in large multimodal models.
\newblock \emph{arXiv preprint arXiv:2407.07895}, 2024{\natexlab{b}}.

\bibitem[Li et~al.(2023)Li, He, Wang, Li, Wang, Luo, Wang, Wang, and Qiao]{li2023videochat}
KunChang Li, Yinan He, Yi Wang, Yizhuo Li, Wenhai Wang, Ping Luo, Yali Wang, Limin Wang, and Yu Qiao.
\newblock Videochat: Chat-centric video understanding.
\newblock \emph{arXiv preprint arXiv:2305.06355}, 2023.

\bibitem[Li et~al.(2024{\natexlab{c}})Li, Wang, He, Li, Wang, Liu, Wang, Xu, Chen, Luo, et~al.]{li2024mvbench}
Kunchang Li, Yali Wang, Yinan He, Yizhuo Li, Yi Wang, Yi Liu, Zun Wang, Jilan Xu, Guo Chen, Ping Luo, et~al.
\newblock Mvbench: A comprehensive multi-modal video understanding benchmark.
\newblock In \emph{Proceedings of the IEEE/CVF Conference on Computer Vision and Pattern Recognition}, pages 22195--22206, 2024{\natexlab{c}}.

\bibitem[Li et~al.(2024{\natexlab{d}})Li, Wang, and Jia]{li2024llama}
Yanwei Li, Chengyao Wang, and Jiaya Jia.
\newblock Llama-vid: An image is worth 2 tokens in large language models.
\newblock In \emph{European Conference on Computer Vision}, pages 323--340. Springer, 2024{\natexlab{d}}.

\bibitem[Lin et~al.(2023)Lin, Ye, Zhu, Cui, Ning, Jin, and Yuan]{lin2023videollava}
Bin Lin, Yang Ye, Bin Zhu, Jiaxi Cui, Munan Ning, Peng Jin, and Li Yuan.
\newblock Video-llava: Learning united visual representation by alignment before projection.
\newblock \emph{arXiv preprint arXiv:2311.10122}, 2023.

\bibitem[Lin et~al.(2024)Lin, Yin, Ping, Molchanov, Shoeybi, and Han]{lin2024vila}
Ji Lin, Hongxu Yin, Wei Ping, Pavlo Molchanov, Mohammad Shoeybi, and Song Han.
\newblock Vila: On pre-training for visual language models.
\newblock In \emph{Proceedings of the IEEE/CVF Conference on Computer Vision and Pattern Recognition}, pages 26689--26699, 2024.

\bibitem[Liu et~al.(2024{\natexlab{a}})Liu, Li, Tang, Ge, Shan, and Li]{liu2024st}
Ruyang Liu, Chen Li, Haoran Tang, Yixiao Ge, Ying Shan, and Ge Li.
\newblock St-llm: Large language models are effective temporal learners.
\newblock In \emph{European Conference on Computer Vision}, pages 1--18. Springer, 2024{\natexlab{a}}.

\bibitem[Liu et~al.(2024{\natexlab{b}})Liu, Cheng, Liu, Zhang, Li, Ren, Zou, Yang, Su, Zhu, et~al.]{liu2024llavaplus}
Shilong Liu, Hao Cheng, Haotian Liu, Hao Zhang, Feng Li, Tianhe Ren, Xueyan Zou, Jianwei Yang, Hang Su, Jun Zhu, et~al.
\newblock Llava-plus: Learning to use tools for creating multimodal agents.
\newblock In \emph{European Conference on Computer Vision}, pages 126--142. Springer, 2024{\natexlab{b}}.

\bibitem[Liu et~al.(2024{\natexlab{c}})Liu, Zhu, Shi, Zhang, Lou, Yang, Xi, Cao, Gu, Li, et~al.]{liu2024nvila}
Zhijian Liu, Ligeng Zhu, Baifeng Shi, Zhuoyang Zhang, Yuming Lou, Shang Yang, Haocheng Xi, Shiyi Cao, Yuxian Gu, Dacheng Li, et~al.
\newblock Nvila: Efficient frontier visual language models.
\newblock \emph{arXiv preprint arXiv:2412.04468}, 2024{\natexlab{c}}.

\bibitem[Maaz et~al.(2023)Maaz, Rasheed, Khan, and Khan]{maaz2023videochatgpt}
Muhammad Maaz, Hanoona Rasheed, Salman Khan, and Fahad~Shahbaz Khan.
\newblock Video-chatgpt: Towards detailed video understanding via large vision and language models.
\newblock \emph{arXiv preprint arXiv:2306.05424}, 2023.

\bibitem[Menon et~al.(2024)Menon, Zemel, and Vondrick]{menon2024whiteboard}
Sachit Menon, Richard Zemel, and Carl Vondrick.
\newblock Whiteboard-of-thought: Thinking step-by-step across modalities.
\newblock \emph{arXiv preprint arXiv:2406.14562}, 2024.

\bibitem[Min et~al.(2024)Min, Buch, Nagrani, Cho, and Schmid]{min2024morevqa}
Juhong Min, Shyamal Buch, Arsha Nagrani, Minsu Cho, and Cordelia Schmid.
\newblock Morevqa: Exploring modular reasoning models for video question answering.
\newblock In \emph{Proceedings of the IEEE/CVF Conference on Computer Vision and Pattern Recognition}, pages 13235--13245, 2024.

\bibitem[Potamias et~al.(2025)Potamias, Zhang, Deng, and Zafeiriou]{potamias2025wilor}
Rolandos~Alexandros Potamias, Jinglei Zhang, Jiankang Deng, and Stefanos Zafeiriou.
\newblock Wilor: End-to-end 3d hand localization and reconstruction in-the-wild.
\newblock In \emph{Proceedings of the Computer Vision and Pattern Recognition Conference}, pages 12242--12254, 2025.

\bibitem[Radford et~al.(2021)Radford, Kim, Hallacy, Ramesh, Goh, Agarwal, Sastry, Askell, Mishkin, Clark, et~al.]{radford2021learning}
Alec Radford, Jong~Wook Kim, Chris Hallacy, Aditya Ramesh, Gabriel Goh, Sandhini Agarwal, Girish Sastry, Amanda Askell, Pamela Mishkin, Jack Clark, et~al.
\newblock Learning transferable visual models from natural language supervision.
\newblock In \emph{International conference on machine learning}, pages 8748--8763. PMLR, 2021.

\bibitem[Ravi et~al.(2024)Ravi, Gabeur, Hu, Hu, Ryali, Ma, Khedr, R{\"a}dle, Rolland, Gustafson, et~al.]{ravi2024sam}
Nikhila Ravi, Valentin Gabeur, Yuan-Ting Hu, Ronghang Hu, Chaitanya Ryali, Tengyu Ma, Haitham Khedr, Roman R{\"a}dle, Chloe Rolland, Laura Gustafson, et~al.
\newblock Sam 2: Segment anything in images and videos.
\newblock \emph{arXiv preprint arXiv:2408.00714}, 2024.

\bibitem[Shang et~al.(2024)Shang, You, Subramanian, Darrell, and Herzig]{shang2024traveler}
Chuyi Shang, Amos You, Sanjay Subramanian, Trevor Darrell, and Roei Herzig.
\newblock Traveler: A modular multi-lmm agent framework for video question-answering.
\newblock \emph{arXiv preprint arXiv:2404.01476}, 2024.

\bibitem[Shtedritski et~al.(2023)Shtedritski, Rupprecht, and Vedaldi]{shtedritski2023does}
Aleksandar Shtedritski, Christian Rupprecht, and Andrea Vedaldi.
\newblock What does clip know about a red circle? visual prompt engineering for vlms.
\newblock In \emph{Proceedings of the IEEE/CVF International Conference on Computer Vision}, pages 11987--11997, 2023.

\bibitem[Sur{\'\i}s et~al.(2023)Sur{\'\i}s, Menon, and Vondrick]{suris2023vipergpt}
D{\'\i}dac Sur{\'\i}s, Sachit Menon, and Carl Vondrick.
\newblock Vipergpt: Visual inference via python execution for reasoning.
\newblock In \emph{Proceedings of the IEEE/CVF International Conference on Computer Vision}, pages 11888--11898, 2023.

\bibitem[Team et~al.(2024)Team, Georgiev, Lei, Burnell, Bai, Gulati, Tanzer, Vincent, Pan, Wang, et~al.]{team2024gemini}
Gemini Team, Petko Georgiev, Ving~Ian Lei, Ryan Burnell, Libin Bai, Anmol Gulati, Garrett Tanzer, Damien Vincent, Zhufeng Pan, Shibo Wang, et~al.
\newblock Gemini 1.5: Unlocking multimodal understanding across millions of tokens of context.
\newblock \emph{arXiv preprint arXiv:2403.05530}, 2024.

\bibitem[Wang et~al.(2024{\natexlab{a}})Wang, Chen, Liu, Chen, Lin, Han, et~al.]{yolo}
Ao Wang, Hui Chen, Lihao Liu, Kai Chen, Zijia Lin, Jungong Han, et~al.
\newblock Yolov10: Real-time end-to-end object detection.
\newblock \emph{Advances in Neural Information Processing Systems}, 37:\penalty0 107984--108011, 2024{\natexlab{a}}.

\bibitem[Wang et~al.(2024{\natexlab{b}})Wang, Wang, Huang, Huang, and Jin]{wang2024videoclip}
Jiapeng Wang, Chengyu Wang, Kunzhe Huang, Jun Huang, and Lianwen Jin.
\newblock Videoclip-xl: Advancing long description understanding for video clip models.
\newblock \emph{arXiv preprint arXiv:2410.00741}, 2024{\natexlab{b}}.

\bibitem[Wang et~al.(2024{\natexlab{c}})Wang, Bai, Tan, Wang, Fan, Bai, Chen, Liu, Wang, Ge, et~al.]{wang2024qwen2}
Peng Wang, Shuai Bai, Sinan Tan, Shijie Wang, Zhihao Fan, Jinze Bai, Keqin Chen, Xuejing Liu, Jialin Wang, Wenbin Ge, et~al.
\newblock Qwen2-vl: Enhancing vision-language model's perception of the world at any resolution.
\newblock \emph{arXiv preprint arXiv:2409.12191}, 2024{\natexlab{c}}.

\bibitem[Wang et~al.(2024{\natexlab{d}})Wang, Zhang, Zohar, and Yeung-Levy]{wang2024videoagent}
Xiaohan Wang, Yuhui Zhang, Orr Zohar, and Serena Yeung-Levy.
\newblock Videoagent: Long-form video understanding with large language model as agent.
\newblock In \emph{European Conference on Computer Vision}, pages 58--76. Springer, 2024{\natexlab{d}}.

\bibitem[Wu et~al.(2023)Wu, Bansal, Zhang, Wu, Zhang, Zhu, Li, Jiang, Zhang, and Wang]{wu2023autogen}
Qingyun Wu, Gagan Bansal, Jieyu Zhang, Yiran Wu, Shaokun Zhang, Erkang Zhu, Beibin Li, Li Jiang, Xiaoyun Zhang, and Chi Wang.
\newblock Autogen: Enabling next-gen llm applications via multi-agent conversation framework.
\newblock \emph{arXiv preprint arXiv:2308.08155}, 2023.

\bibitem[Wu et~al.(2025)Wu, Mao, Zhang, Xia, Dong, Cui, and Wei]{wu2025mind}
Wenshan Wu, Shaoguang Mao, Yadong Zhang, Yan Xia, Li Dong, Lei Cui, and Furu Wei.
\newblock Mind's eye of llms: visualization-of-thought elicits spatial reasoning in large language models.
\newblock \emph{Advances in Neural Information Processing Systems}, 37:\penalty0 90277--90317, 2025.

\bibitem[Wu et~al.(2024)Wu, Wang, Tang, Wu, He, Ouyang, Torr, and Wu]{wu2024dettoolchain}
Yixuan Wu, Yizhou Wang, Shixiang Tang, Wenhao Wu, Tong He, Wanli Ouyang, Philip Torr, and Jian Wu.
\newblock Dettoolchain: A new prompting paradigm to unleash detection ability of mllm.
\newblock In \emph{European Conference on Computer Vision}, pages 164--182. Springer, 2024.

\bibitem[Xu et~al.(2024)Xu, Zhao, Zhou, Lin, Ng, and Feng]{xu2024pllava}
Lin Xu, Yilin Zhao, Daquan Zhou, Zhijie Lin, See~Kiong Ng, and Jiashi Feng.
\newblock Pllava: Parameter-free llava extension from images to videos for video dense captioning.
\newblock \emph{arXiv preprint arXiv:2404.16994}, 2024.

\bibitem[Xue et~al.(2024)Xue, Chen, Li, Hu, Zhu, Li, Fang, Tang, Yang, Liu, et~al.]{xue2024longvila}
Fuzhao Xue, Yukang Chen, Dacheng Li, Qinghao Hu, Ligeng Zhu, Xiuyu Li, Yunhao Fang, Haotian Tang, Shang Yang, Zhijian Liu, et~al.
\newblock Longvila: Scaling long-context visual language models for long videos.
\newblock \emph{arXiv preprint arXiv:2408.10188}, 2024.

\bibitem[Yang et~al.(2024{\natexlab{a}})Yang, Huang, Lu, Han, Zhang, Gao, Hu, and Zhao]{yang2024vript}
Dongjie Yang, Suyuan Huang, Chengqiang Lu, Xiaodong Han, Haoxin Zhang, Yan Gao, Yao Hu, and Hai Zhao.
\newblock Vript: A video is worth thousands of words.
\newblock \emph{arXiv preprint arXiv:2406.06040}, 2024{\natexlab{a}}.

\bibitem[Yang et~al.(2023{\natexlab{a}})Yang, Zhang, Li, Zou, Li, and Gao]{yang2023set}
Jianwei Yang, Hao Zhang, Feng Li, Xueyan Zou, Chunyuan Li, and Jianfeng Gao.
\newblock Set-of-mark prompting unleashes extraordinary visual grounding in gpt-4v.
\newblock \emph{arXiv preprint arXiv:2310.11441}, 2023{\natexlab{a}}.

\bibitem[Yang et~al.(2023{\natexlab{b}})Yang, Li, Wang, Lin, Azarnasab, Ahmed, Liu, Liu, Zeng, and Wang]{yang2023mmreact}
Zhengyuan Yang, Linjie Li, Jianfeng Wang, Kevin Lin, Ehsan Azarnasab, Faisal Ahmed, Zicheng Liu, Ce Liu, Michael Zeng, and Lijuan Wang.
\newblock Mm-react: Prompting chatgpt for multimodal reasoning and action.
\newblock \emph{arXiv preprint arXiv:2303.11381}, 2023{\natexlab{b}}.

\bibitem[Yang et~al.(2024{\natexlab{b}})Yang, Chen, Li, Wang, and Yang]{yang2024doraemongpt}
Zongxin Yang, Guikun Chen, Xiaodi Li, Wenguan Wang, and Yi Yang.
\newblock Doraemongpt: Toward understanding dynamic scenes with large language models (exemplified as a video agent).
\newblock In \emph{Forty-first International Conference on Machine Learning}, 2024{\natexlab{b}}.

\bibitem[Ye et~al.(2024)Ye, Xu, Liu, Hu, Yan, Qian, Zhang, Huang, and Zhou]{ye2024mplugowl3}
Jiabo Ye, Haiyang Xu, Haowei Liu, Anwen Hu, Ming Yan, Qi Qian, Ji Zhang, Fei Huang, and Jingren Zhou.
\newblock mplug-owl3: Towards long image-sequence understanding in multi-modal large language models.
\newblock In \emph{The Thirteenth International Conference on Learning Representations}, 2024.

\bibitem[Zhang et~al.(2025{\natexlab{a}})Zhang, Li, Cheng, Hu, Yuan, Chen, Leng, Jiang, Zhang, Li, et~al.]{zhang2025videollama3}
Boqiang Zhang, Kehan Li, Zesen Cheng, Zhiqiang Hu, Yuqian Yuan, Guanzheng Chen, Sicong Leng, Yuming Jiang, Hang Zhang, Xin Li, et~al.
\newblock Videollama 3: Frontier multimodal foundation models for image and video understanding.
\newblock \emph{arXiv preprint arXiv:2501.13106}, 2025{\natexlab{a}}.

\bibitem[Zhang et~al.(2023)Zhang, Li, and Bing]{zhang2023videollama}
Hang Zhang, Xin Li, and Lidong Bing.
\newblock Video-llama: An instruction-tuned audio-visual language model for video understanding.
\newblock \emph{arXiv preprint arXiv:2306.02858}, 2023.

\bibitem[Zhang et~al.(2025{\natexlab{b}})Zhang, Deng, Ma, and Potamias]{zhang2025hawor}
Jinglei Zhang, Jiankang Deng, Chao Ma, and Rolandos~Alexandros Potamias.
\newblock Hawor: World-space hand motion reconstruction from egocentric videos.
\newblock In \emph{Proceedings of the Computer Vision and Pattern Recognition Conference (CVPR)}, pages 1805--1815, 2025{\natexlab{b}}.

\bibitem[Zhang et~al.(2024)Zhang, Wu, Li, Li, Ma, Liu, and Li]{zhang2024llavavideo}
Yuanhan Zhang, Jinming Wu, Wei Li, Bo Li, Zejun Ma, Ziwei Liu, and Chunyuan Li.
\newblock Video instruction tuning with synthetic data.
\newblock \emph{arXiv preprint arXiv:2410.02713}, 2024.

\end{thebibliography}
}

\end{document}